\pdfoutput=1

\documentclass[11pt]{article}

\usepackage[final]{acl}

\usepackage{times}
\usepackage{latexsym}

\usepackage{xurl}

\usepackage[T1]{fontenc}

\usepackage[utf8]{inputenc}

\usepackage{microtype}

\usepackage{inconsolata}

\usepackage{graphicx}

\title{Selective Token Pruning for Efficient Korean Large Language Models}

\author{First Author \\
  Affiliation / Address line 1 \\
  Affiliation / Address line 2 \\
  Affiliation / Address line 3 \\
  \texttt{email@domain} \\\And
  Second Author \\
  Affiliation / Address line 1 \\
  Affiliation / Address line 2 \\
  Affiliation / Address line 3 \\
  \texttt{email@domain} \\}

\usepackage{amsmath}
\usepackage{graphicx}
\usepackage{booktabs}
\usepackage{multirow}
\title{Optimizing Korean-Centric LLMs via Token Pruning}
\author{Hoyeol Kim \\
  College of Computing \\
  Georgia Institute of Technology \\
  \texttt{hkim3263@gatech.edu} \\\And
  Hyeonwoo Kim \\
  Independent Researcher \\
  \texttt{kimhyeonwoo2431@gmail.com} \\}
\date{}
\begin{document}
\maketitle

\begin{abstract}
This paper presents a systematic benchmark of state-of-the-art multilingual large language models (LLMs) adapted via token pruning—a compression technique that eliminates tokens and embedding parameters corresponding to languages irrelevant to the target application. Focusing on Korean-centric natural language processing (NLP) tasks, we evaluate architectures including Qwen3, Gemma-3, Llama-3, and Aya across three vocabulary configurations: Original, English-Korean (EnKo), and English-Korean-Chinese (EnKoZh). Performance is assessed using established benchmarks for general aptitude, cultural literacy, instruction following, and machine translation. Our findings indicate that token pruning significantly improves generation stability by eliminating language confusion, and in the case of machine translation, frequently enhances performance on Korean-specific tasks. While instruction-following capabilities display architecture-dependent variance linked to latent cross-lingual representations, the significant reduction in vocabulary size validates token pruning as a highly effective optimization strategy for memory-constrained, domain-specific deployments, despite modest gains in inference latency.
\end{abstract}

\section{Introduction}
The proliferation of massively multilingual LLMs has significantly expanded the accessibility of advanced natural language processing. Models such as LLaMA-3 \citep{dubey2024llama}, Qwen \citep{yang2025qwen3}, and Gemma \citep{team2025gemma} rely on vast multilingual training corpora to achieve universal applicability. However, this breadth introduces a "curse of multilinguality," wherein a substantial fraction of the model's parameters—particularly within the vocabulary and embedding layers—is allocated to languages that are superfluous for specific downstream applications. For deployments in strictly monolingual or bilingual contexts, such as Korean-centric services, this redundancy imposes unnecessary memory overhead and computational inefficiency.

In the context of Korean NLP, this challenge is two-pronged. First, despite being a high-resource language, Korean often comprises a small minority of the training data in dominant English- or Chinese-centric models, which risks diluting cultural nuance. Second, while local deployment requires lightweight models, optimization strategies must not come at the cost of linguistic proficiency.

Token pruning offers a targeted solution. By systematically excising tokens linked to non-target languages, this method compresses the vocabulary and embedding matrices, reducing the memory footprint without altering the core transformer architecture. While prior works have explored pruning generally, systematic research quantifying its effects on Korean language tasks remains limited. This paper bridges that gap by benchmarking token pruning on State-of-the-Art (SOTA) multilingual LLMs. We specifically evaluate how aggressive vocabulary compression impacts performance across three distinct pillars: general aptitude, cultural competence, and translation quality.

\section{Related Work}

\subsection{Token Pruning}
Token pruning is increasingly recognized as a vital component of transformer efficiency. \citet{accelerating2025survey}  provide a comprehensive review of token pruning methods, categorizing them into heuristic-based, learnable, and reinforcement learning-based approaches, and emphasize its role in reducing inference complexity while retaining accuracy. In multimodal contexts, \citet{wen2024multimodal} question the utility of complex attention-based importance scores, demonstrating that simple random baselines can approach the performance of specialized methods that lack strong theoretical grounding. Building on the theoretical view of token redundancy in transformers, \citet{tai2025role} formalized benchmarks to capture FLOP reductions and latency trade-offs under different pruning strategies. Complementing these input-level approaches, \citet{lee2024dynamic} introduced dynamic vocabulary pruning to restrict the output search space during testing, suggesting a synergistic effect between vocabulary- and token-level pruning for practical deployment.

\subsection{Korean-Centric LLMs and Language Specialization}
Efforts to enhance Korean LLM performance bifurcate into adapting English-centric models and developing sovereign AI. \citet{kim2024adaptedllm} showed that continual pretraining with tokenizer augmentation effectively adapts English foundation models, while \citet{kim2025thunderllm} introduced Thunder-LLM, achieving near-SOTA results via balanced bilingual corpora. Domain-specific applications have also expanded, exemplified by the bilingual GECKO model for code generation \citep{oh2018gecko} and the FINKRX model for financial NLP \citep{son2025finkrx}.


\section{Methods}

\subsection{Target Models and Configurations}
We evaluate a diverse set of open-weight multilingual LLMs, specifically the Qwen3 series (ranging from 0.6B to 14B parameters), Gemma-3 (270M to 12B), and the Llama-3 family (3.1-8B and 3.2-3B). Additional models include Tri \citep{trillionlabs_tri7b}, Ministral-8B \citep{liu2026ministral}, and Aya-23 \citep{aryabumi2024aya}. To assess the impact of vocabulary compression, each model is processed into three distinct configurations: the Original setup, which retains the full multilingual vocabulary; the EnKo configuration, pruned to retain only English and Korean tokens; and the EnKoZh configuration, which preserves English, Korean, and Chinese tokens. The pruning process necessitates tokenizer reconstruction, positional encoding realignment, and the remapping of output layer weights to correspond to the reduced vocabulary indices.

\subsection{Vocabulary Pruning Procedure}
We implement a ``language-aware filtering'' strategy that proceeds in three distinct stages. First, tokens are categorized by language using Unicode block ranges and script properties such as Hangul, Latin, and Hanzi (Chinese characters). Subsequently, tokens irrelevant to the target configuration—whether EnKo or EnKoZh—are discarded, and the retained tokens are mapped onto a new, continuous index space. Finally, the embedding matrix and output projection layer are physically rearranged to align with the new indices, effectively reducing the parameter count. This procedure leaves the internal transformer blocks and positional encodings intact, preserving the model's sequence modeling capabilities.

\subsection{Benchmarks and Metrics}
To quantify performance, we utilize four categories of Korean-centric benchmarks. General aptitude is assessed via KMMLU \citep{son2025kmmlu}, which tests general knowledge and logical reasoning. Cultural and linguistic understanding are evaluated using HAERAE \citep{son2024hae} for cultural literacy
and CLIcK \citep{kim2024click} for linguistic nuance. Instruction-following capabilities are measured through LogicKor \citep{logickor} and KoMTBench \citep{KoMT-Bench}, which assess reasoning and constraint satisfaction. Finally, machine translation performance is evaluated using the WMT 24++ (Korean--English) benchmark \citep{deutsch2025wmt24++}, scored via XCOMET-XXL \citep{guerreiro2024xcomet}. 

\subsection{Latency Measurement} 
Token-level inference latency is quantified using the Seed-X-PPO-7B model \citep{cheng2025seedxbuildingstrongmultilingual}
on the Google/WMT-pp dataset \citep{deutsch2025wmt24++}. We compare configurations under identical hardware conditions to isolate the speedups resulting from the reduced softmax computational load.

\section{Experiments and Analysis}

\begin{table*}[t!]
\centering
\small
\caption{General Aptitude (KMMLU), Cultural (HAERAE), and Linguistic (CLIcK) Benchmark Results. Scores: accuracy. Higher values are more accurate.}
\label{tab:gen_culture}
\resizebox{\textwidth}{!}{%
\begin{tabular}{l|ccc|ccc|ccc}
\toprule
\multirow{2}{*}{\textbf{Model}} & \multicolumn{3}{c|}{\textbf{KMMLU}} & \multicolumn{3}{c|}{\textbf{HAERAE}} & \multicolumn{3}{c}{\textbf{CLIcK}} \\
\cmidrule{2-10}
& \textbf{Original} & \textbf{EnKo} & \textbf{EnKoZh} & \textbf{Original} & \textbf{EnKo} & \textbf{EnKoZh} & \textbf{Original} & \textbf{EnKo} & \textbf{EnKoZh} \\
\midrule
Qwen3-0.6B & 0.1482 & 0.1482 & 0.1483 & 0.2328 & 0.2328 & 0.2328 & 0.3188 & 0.3188 & 0.3183 \\
Qwen3-1.7B & 0.3589 & 0.3589 & 0.3591 & 0.4244 & 0.4235 & 0.4235 & 0.4010 & 0.3980 & 0.4005 \\
Qwen3-4B & 0.4581 & 0.4583 & 0.4585 & 0.5243 & 0.5234 & 0.5234 & 0.5574 & 0.5554 & 0.5569 \\
Qwen3-8B & 0.5254 & 0.5244 & 0.5249 & 0.5995 & 0.5995 & 0.5995 & 0.6005 & 0.5985 & 0.6005 \\
Qwen3-14B & 0.5340 & 0.5320 & 0.5325 & 0.6700 & 0.6700 & 0.6700 & 0.6221 & 0.6206 & 0.6221 \\
\midrule
Gemma-3-270m-it & 0.3057 & 0.3058 & 0.3056 & 0.1989 & 0.1989 & 0.1989 & 0.3033 & 0.3033 & 0.3033 \\
Gemma-3-1b-it & 0.3094 & 0.3086 & 0.3090 & 0.3575 & 0.3584 & 0.3584 & 0.3388 & 0.3373 & 0.3378 \\
Gemma-3-4b-it & 0.3801 & 0.3788 & 0.3792 & 0.5170 & 0.5170 & 0.5160 & 0.4982 & 0.4982 & 0.4997 \\
Gemma-3-12b-it & 0.4812 & 0.4804 & 0.4806 & 0.6838 & 0.6838 & 0.6838 & 0.6311 & 0.6321 & 0.6321 \\
\midrule
Llama-3.1-8B-Inst & 0.4165 & 0.4153 & 0.4156 & 0.5775 & 0.5775 & 0.5775 & 0.5213 & 0.5178 & 0.5208 \\
Llama-3.2-3B-Inst & 0.3307 & 0.3311 & 0.3311 & 0.3639 & 0.3621 & 0.3621 & 0.4190 & 0.4190 & 0.4180 \\
Tri-7B & 0.4944 & 0.4932 & 0.4934 & 0.7929 & 0.7919 & 0.7929 & 0.6486 & 0.6471 & 0.6486 \\
Aya-23-8B & 0.1788 & 0.1784 & 0.1787 & 0.5619 & 0.5619 & 0.5619 & 0.4802 & 0.4802 & 0.4792 \\
Aya-expanse-8b & 0.3436 & 0.3427 & 0.3428 & 0.6627 & 0.6627 & 0.6636 & 0.5153 & 0.5138 & 0.5153 \\
\bottomrule
\end{tabular}%
}
\end{table*}

\subsection{General Aptitude and Cultural Alignment}
Table \ref{tab:gen_culture} summarizes the assessment of fundamental Korean language grasp and cultural knowledge. The data reveals minimal performance degradation in pruned models. For the Qwen3 series, the variance between configurations is negligible ($<0.01$ fluctuation), while Gemma-3-12b-it shows a slight improvement in linguistic capability ($0.6311 \rightarrow 0.6321$). This stability suggests that pruning successfully eliminates vocabulary redundancy without excising the semantic structures requisite for Korean reasoning. Furthermore, Llama-3.2-3B-Inst exhibits a marginal increase in KMMLU accuracy after pruning ($0.3307 \rightarrow 0.3311$), implying that narrowing the vocabulary search space may reduce generation noise in lower-parameter models.

\subsection{Instruction-Following Capabilities}
Performance in complex instruction following (LogicKor and KoMTBench) exhibits greater variance, influenced by the interaction between model architecture and pruning depth (Table \ref{tab:instruction}). For the Qwen3 family, retaining Chinese tokens (EnKoZh) consistently outperforms the stricter EnKo pruning. Notably, Qwen3-4B scores $7.85$ in LogicKor (EnKoZh), surpassing its Original score ($7.77$). This suggests that models heavily pre-trained on Chinese corpora rely on latent cross-lingual alignments for reasoning. Conversely, larger models like Llama-3.1-8B-Inst show improved performance in the EnKo setting ($5.36 \rightarrow 5.57$), indicating that for sufficiently large models, vocabulary reduction may sharpen instruction focus.

\begin{table}[h!]
\centering
\small
\caption{Instruction-Following Results (LogicKor, KoMTBench). Scores: higher is better.}
\label{tab:instruction}
\resizebox{\columnwidth}{!}{%
\begin{tabular}{l|ccc|ccc}
\toprule
\multirow{2}{*}{\textbf{Model}} & \multicolumn{3}{c|}{\textbf{LogicKor}} & \multicolumn{3}{c}{\textbf{KoMTBench}} \\
\cmidrule{2-7}
& \textbf{Original} & \textbf{EnKo} & \textbf{EnKoZh} & \textbf{Original} & \textbf{EnKo} & \textbf{EnKoZh} \\
\midrule
Qwen3-0.6B & 3.43 & 3.11 & 3.32 & 3.550 & 3.516 & 3.558 \\
Qwen3-1.7B & 5.92 & 5.61 & 5.71 & 6.431 & 5.588 & 5.516 \\
Qwen3-4B & 7.77 & 7.44 & 7.85 & 7.822 & 7.824 & 7.971 \\
Qwen3-8B & 8.71 & 8.33 & 8.52 & 8.429 & 7.713 & 7.857 \\
Qwen3-14B & 8.71 & 8.87 & 8.85 & 8.496 & 8.673 & 8.564 \\
\midrule
Gemma-3-270m-it & 1.62 & 1.57 & 1.43 & 1.871 & 1.833 & 2.015 \\
Gemma-3-1b-it & 5.04 & 5.05 & 4.77 & 5.874 & 5.927 & 5.874 \\
Gemma-3-4b-it & 7.86 & 8.12 & 8.20 & 8.294 & 8.213 & 8.358 \\
Gemma-3-12b-it & 8.89 & 8.93 & 9.18 & 8.635 & 8.408 & 8.485 \\
\midrule
Llama-3.1-8B-Inst & 5.36 & 5.57 & 5.55 & 5.820 & 5.816 & 5.673 \\
Llama-3.2-3B-Inst & 3.11 & 3.31 & 3.32 & 2.978 & 3.130 & 3.118 \\
Tri-7B & 8.46 & 8.02 & 8.27 & 8.370 & 8.060 & 8.544 \\
Aya-23-8B & 6.18 & 6.14 & 5.92 & 5.404 & 5.564 & 5.553 \\
Aya-expanse-8b & 7.87 & 7.73 & 7.64 & 7.547 & 7.368 & 7.579 \\
\bottomrule
\end{tabular}%
}
\end{table}

\subsection{Machine Translation}
Machine translation (WMT24++) results, presented in Table \ref{tab:translation}, provide the strongest evidence for the efficacy of token pruning. Pruned configurations consistently match or outperform baselines. Llama-3.1-8B-Inst ($0.5879 \rightarrow 0.6342$) and Aya-expanse-8b ($0.6957 \rightarrow 0.7496$) show substantial gains. This indicates that eliminating extraneous language tokens regularizes the output distribution, minimizing off-target hallucinations and enhancing the English-Korean translation pathway.

\begin{table}[h!]
\centering
\small
\caption{Machine Translation Evaluation Results (WMT24 Korean–English). Scores: COMET. Higher is better.}
\label{tab:translation}
\resizebox{\columnwidth}{!}{%
\begin{tabular}{lccc}
\toprule
\textbf{Model} & \textbf{Original} & \textbf{EnKo} & \textbf{EnKoZh} \\
\midrule
Qwen3-0.6B & 0.4014 & 0.4025 & 0.4053 \\
Qwen3-1.7B & 0.5465 & 0.5489 & 0.5489 \\
Qwen3-4B & 0.6611 & 0.6613 & 0.6585 \\
Qwen3-8B & 0.7296 & 0.7347 & 0.7320 \\
Qwen3-14B & 0.7612 & 0.7650 & 0.7635 \\
\midrule
Gemma-3-270m-it & 0.4698 & 0.4742 & 0.4725 \\
Gemma-3-1b-it & 0.1426 & 0.1474 & 0.1423 \\
Gemma-3-4b-it & 0.1208 & 0.1224 & 0.1217 \\
Gemma-3-12b-it & 0.1424 & 0.1428 & 0.1416 \\
\midrule
Llama-3.1-8B-Inst & 0.5879 & 0.6342 & 0.6326 \\
Llama-3.2-3B-Inst & 0.3081 & 0.3080 & 0.3078 \\
Tri-7B & 0.4081 & 0.4627 & 0.4632 \\
Tri-1.8B-Translation & 0.6202 & 0.6177 & 0.6190 \\
Aya-23-8B & 0.7247 & 0.7338 & 0.7338 \\
Aya-expanse-8b & 0.6957 & 0.7496 & 0.7496 \\
Seed-X-PPO-7B & 0.7679 & 0.7719 & 0.7705 \\
\bottomrule
\end{tabular}%
}
\end{table}

\subsection{Language Confusion Performance}
We measured the Word-level Pass Rate (WPR) to evaluate stability in language generation,
following the methodology of Marchisio et al. \citep{marchisio2024understanding}. Table \ref{tab:wpr_delta} details the improvements yielded by the EnKo adaptation. The EnKo method significantly mitigates language confusion. \textsc{Qwen3-4B}, which exhibited the lowest baseline stability ($0.8882$), demonstrated the most dramatic recovery ($\Delta\text{WPR} = +0.1041$). Even stable baselines like \textsc{Qwen3-0.6B} achieved near-perfect consistency ($>0.999$) post-pruning.

\begin{table}[h!]
\centering
\footnotesize
\setlength{\tabcolsep}{2.5pt}
\caption{Word-level Pass Rate (WPR) Improvement by EnKo}
\label{tab:wpr_delta}
\resizebox{\columnwidth}{!}{%
\begin{tabular}{lccc}
\hline
\textbf{Model} & \textbf{Base WPR} & \textbf{EnKo WPR} & $\Delta$ \textbf{WPR} \\ \hline
Qwen3-0.6B            & 0.9867 & 0.9999 & +0.0132 \\
Qwen3-1.7B            & 0.9622 & 0.9991 & +0.0369 \\
Qwen3-4B              & 0.8882 & 0.9923 & +0.1041 \\
Llama-3.2-3B-Instruct & 0.9704 & 0.9980 & +0.0276 \\
Llama-3.1-8B-Instruct & 0.9765 & 0.9951 & +0.0186 \\ \hline
\end{tabular}%
}
\end{table}

\subsection{Computational Latency}
Efficiency gains were measured using the Seed-X-PPO-7B model (Table \ref{tab:latency}). While vocabulary reduction is substantial (36\% reduction in EnKo), latency improvement is modest (0.89\%). This confirms that while pruning alleviates memory overhead related to embeddings, it does not significantly impact the computational bottlenecks in attention mechanisms.

\begin{table}[h!]
\centering
\caption{Latency comparison for Seed-X-PPO-7B. Measured on Google/WMT-pp.}
\label{tab:latency}
\resizebox{\columnwidth}{!}{%
\begin{tabular}{lcccc}
\toprule
\textbf{Model Config.} & \textbf{Vocab Size} & \textbf{\% of Original} & \textbf{Latency (ms)} & \textbf{Improvement} \\
\midrule
\textbf{Original}     & 65,269 & 100\%  & 1126 ms          & - \\
\textbf{EKC (enkoch)} & 56,660 & 86.8\% & 1122 ms          & 0.36\% \\
\textbf{EK (enko)}    & 41,704 & 63.9\% & \textbf{1116 ms} & \textbf{0.89\%} \\
\bottomrule
\end{tabular}%
}
\end{table}

\section{Conclusion}
We presented a systematic benchmark of token pruning for Korean NLP, confirming that high-resource languages can be effectively decoupled from massive multilingual vocabularies without performance penalty. Pruning eliminated extraneous tokens, leading to near-perfect generation consistency (WPR $>0.99$) and improved translation quality, all while significantly reducing parameter count.

However, our results also highlight critical architectural dependencies. While pruning is generally robust, the performance gap between EnKo and EnKoZh configurations in Qwen series models suggests that latent cross-lingual representations (specifically Chinese) remain integral to reasoning capabilities in certain architectures. Finally, while the inference latency gains were modest ($<1\%$), the substantial reduction in vocabulary size offers significant memory savings. This positions token pruning not merely as a compression technique, but as a vital tool for deploying stable, high-performance sovereign AI models in resource-constrained local environments.


\bibliographystyle{acl_natbib}
\bibliography{custom}

\end{document}